\documentclass[letterpaper, 10 pt, conference]{ieeeconf}  
\let\labelindent\relax

\IEEEoverridecommandlockouts                              

\title{\LARGE \bf
Scaling Whole-body Multi-contact Manipulation \\
with Contact Optimization
}

\author{
    Victor Levé$^{1*}$, João Moura$^{1}$, Sachiya Fujita$^{2}$, Tamon Miyake$^{2}$, Steve Tonneau$^{1}$ and Sethu Vijayakumar$^{1}$ %
    \thanks{$^{1}$ School of Informatics, The University of Edinburgh, UK}%
    \thanks{$^{2}$ Waseda University, JAPAN}%
    \thanks{This work is supported by the JST Moonshot R\&D (Grant No. JPMJMS2031), Kawada Robotics Corporation and The Alan Turing Institute.}%
}

\usepackage{tikz}
\usetikzlibrary{shapes.geometric, arrows, fit, backgrounds}
\usetikzlibrary {arrows.meta,bending}
\usepackage{subcaption}
\usepackage{amsmath}
\usepackage{multirow}
\usepackage[abbreviations]{glossaries-extra}
\usepackage[shortlabels]{enumitem}
\usepackage[utf8]{inputenc}
\usepackage{graphicx}
\usepackage{algorithm}
\usepackage{algpseudocode}
\usepackage{hyperref}
\usepackage[capitalise]{cleveref}

\DeclareMathOperator*{\argmin}{argmin}

\providecommand{\DIFdel}[1]{} 

\newabbreviation{to}{TO}{Trajectory Optimization}
\newabbreviation{co}{CO}{Contact Optimization}
\newabbreviation{rrt}{RRT}{Rapidly-exploring Random Tree}
\newabbreviation{mpc}{MPC}{Model Predictive Control}
\newabbreviation{wmm}{WMM}{Whole-body Multi-contact Manipulation}
\newabbreviation{mppi}{MPPI}{Model Predictive Path Integral}
\newabbreviation{nlp}{NLP}{Nonlinear Programming Problem}
\newabbreviation{dof}{DOF}{Degrees of Freedom}
\newabbreviation{sdf}{SDF}{Signed Distance Field}

\tikzstyle{bloc} = [rectangle, rounded corners, minimum width=2.0cm, minimum height=2.0cm, text centered, draw=black, fill=yellow!10, font=\footnotesize, align=center, anchor=north]
\tikzstyle{circ} = [circle, minimum width=2.0cm, minimum height=2.0cm, text centered, draw=black, fill=yellow!10, font=\footnotesize, align=center, anchor=north]
\tikzstyle{decision} = [diamond, minimum width=2cm, minimum height=1cm, text centered, draw=black, fill=lime!10]
\tikzstyle{arrow} = [thick,->,>=stealth]
\tikzstyle{inout} = [align=center, node distance=1.5cm]

\begin{document}

\maketitle
\global\csname @topnum\endcsname 0
\global\csname @botnum\endcsname 0
\thispagestyle{empty}
\pagestyle{empty}

\begin{abstract}
Daily tasks require us to use our whole body to manipulate objects, for instance when our hands are unavailable. 
We consider the issue of providing humanoid robots with the ability to autonomously perform similar whole-body manipulation tasks. In this context, the infinite possibilities for where and how contact can occur on the robot and object surfaces hinder the scalability of existing planning methods, which predominantly rely on discrete sampling.
Given the continuous nature of contact surfaces, gradient-based optimization offers a more suitable approach for finding solutions. However, a key remaining challenge is the lack of an efficient representation of robot surfaces.
In this work, we propose (i) a representation of robot and object surfaces that enables closed-form computation of proximity points, and (ii) a cost design that effectively guides whole-body manipulation planning.
Our experiments demonstrate that the proposed framework can solve problems unaddressed by existing methods, and achieves a 77\% improvement in planning time over the state of the art. We also validate the suitability of our approach on real hardware through the whole-body manipulation of boxes by a humanoid robot. \href{https://levevictor.github.io/thelazyrobot/#humanoids2025}{Project page}.
\end{abstract}

\section{INTRODUCTION}
\label{section:introduction}

\subsection{Overview}
\begin{figure}[t]
    \centering
    \begin{minipage}{.22\textwidth}
        \begin{subfigure}{\textwidth}
            \includegraphics[width=\textwidth,trim={0 8cm 0 0.6cm},clip]
                {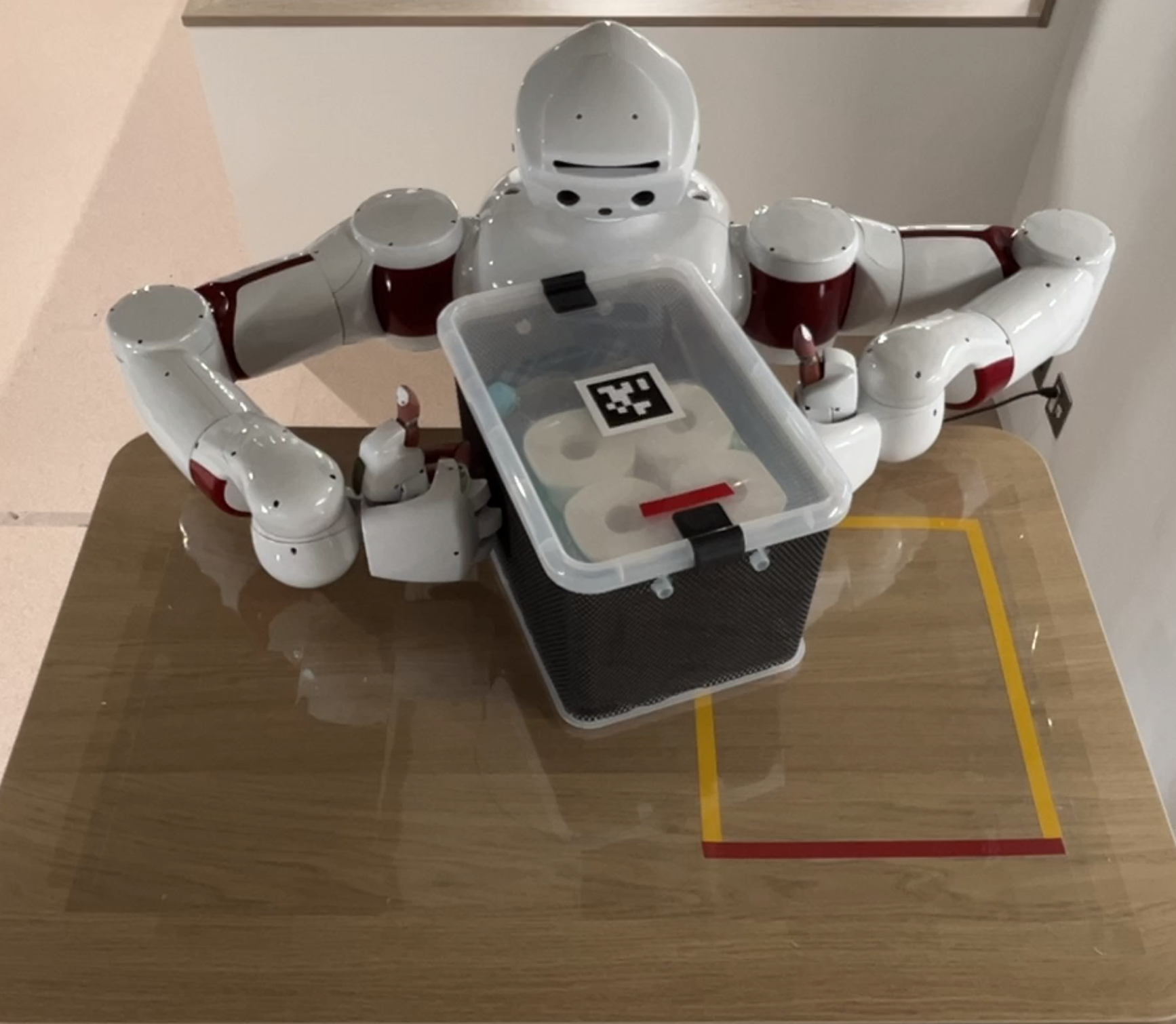}
        \end{subfigure}
        \begin{subfigure}{\textwidth}
            \includegraphics[width=\textwidth,trim={0 0.2cm 0 0.6cm},clip]
                {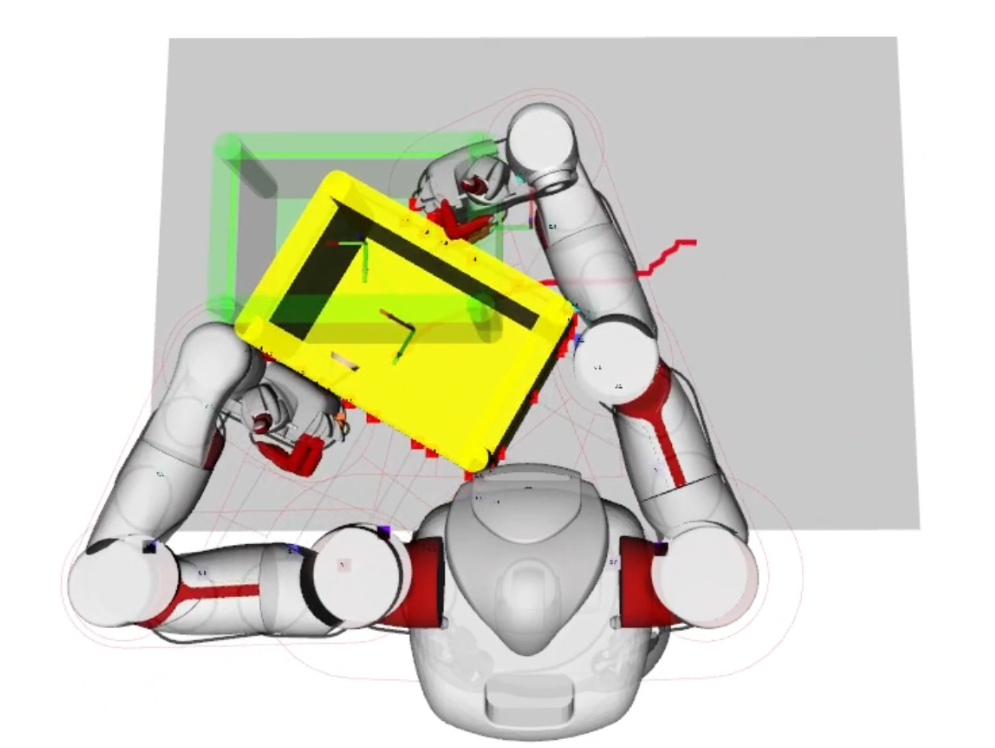}
            \subcaption{Whole-body Manipulation}
            \label{fig:intro_wmm}
        \end{subfigure}
    \end{minipage}
    \begin{minipage}{.25\textwidth}
        \begin{subfigure}{\textwidth}
            \includegraphics[width=\textwidth,trim={4cm 0 0.3cm 0},clip]
                {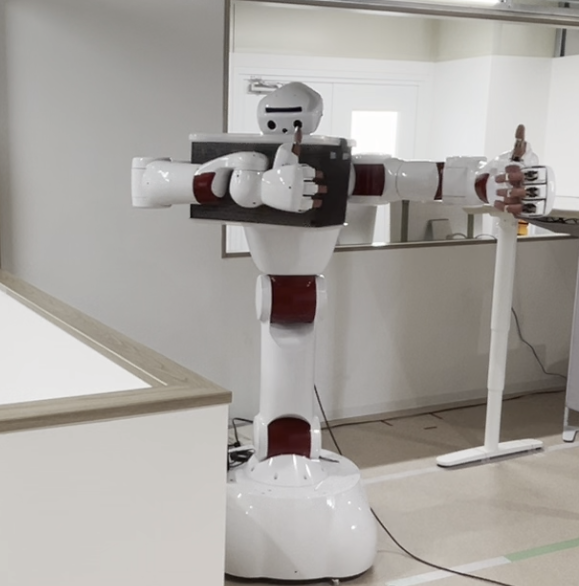}
            \subcaption{Loco-manipulation}
            \label{fig:intro_locom}
        \end{subfigure}
    \end{minipage}
    \centering
    \captionsetup{justification=justified,margin=0.0cm}
    \caption{\gls{wmm} enables versatile handling of objects with the full surface of robot body.}
    \label{fig:intro}
\end{figure}

The autonomy of humanoid robots lies in their ability to automatically translate high-level commands into actionable plans.
In an object manipulation task, a human should only need to specify the desired pose of the object, with the corresponding sequence of contacts and joint trajectory computed autonomously to accomplish the task.
The task objective, along with the geometry and physical properties of both the robot and the object, will dictate different contact strategies.
These may for instance involve using the robot's torso as a contact surface to manipulate an item (\cref{fig:intro_wmm}) or keeping a hand free to open a door (\cref{fig:intro_locom}).
We refer to the problem of computing such motions by leveraging the robot’s full contact surface as the \gls{wmm} problem.

\subsection{Challenges}

Solving a \gls{wmm} problem involves exploring a high-dimensional space to find feasible motions that lie on non-convex contact manifolds, within a reasonable time. 
Unfortunately, planning contact-rich motions is a high-dimensional problem, further complicated by the need to make discrete choices of contact modes (sticking, sliding, breaking) across infinitely many potential contact locations resulting in combinatorial complexity that grows exponentially \cite{Lev__2024}. 

Without an efficient exploration strategy, a \gls{wmm} solver may take hours to find a feasible plan. 
Ideally, a system model would guide the search toward actions that directly support the manipulation goal. 
Given the continuous nature of contact surfaces, this search guide could be enabled with the formulation of the \gls{wmm} planning as a continuous \gls{nlp}.
However, to actualize this, two modeling challenges remain: 
\begin{enumerate}[(a)]
    \item Finding a representation of the object surface and the non-convex robot surface that allows a differentiable computation of the proximity points on these surfaces integrable with a \gls{nlp};
    \item Finding a decision strategy to guide efficiently the planner to avoids local minima, permits manipulation using contact on body parts with lower \gls{dof} with an undefined number of contacts.
\end{enumerate}

To deal with these challenges, related works have relied mainly on discrete sampling to break down the complexity of infinite possibilities of contact location on the continuous surface of the robot \cite{pang2023globalplanningcontactrichmanipulation, jin2024complementarity}. 
However it is unrealistic to use discrete sampling to cover large surfaces like the robot body.
Existing continuous representations of surfaces still require human to predefine which body part will contact with the object \cite{7487300} and are unsuitable for integrating into a whole-body contact planner. In \cite{mordatch2012}, Mordatch et al. formulated a way to solve multi-body contact manipulation with continuous optimization but eluded important contact modes like sliding which prevents its use on body parts with lower \gls{dof} such as the upper arm of the robot.

\subsection{Contribution}
To address the representation problem (a) and the decision problem (b), our contribution is: 
\begin{enumerate}[(i)]
    \item A novel representation of the robot surface suitable for planar \gls{wmm}. 
    This representation is continuous and differentiable and enables formulating the proximity points between the robot and object surfaces as a closed-form solution, easing its integration in \gls{nlp}
    \item A cost design that combines robot-centric and object-centric manipulability metrics with a contact activation function.
    This cost design helps the planner to find grasps with better manipulability and thus less prone to local minima of \gls{wmm}.
    It also permits autonomous planning of manipulations including body parts with lower \gls{dof} and switching between manipulations with a single arm, with dual-arm and with the static torso.
\end{enumerate}
Our experiments show that our representation is accurate enough to permit transfer to real robot hardware with complex geometry. 
We demonstrate the time performance superiority of our planning, with a $77\%$ improvement of the planning time on average and higher success rates over the state of the art \gls{wmm} sampling-based planner \cite{pang2023globalplanningcontactrichmanipulation}. 
We also highlight the benefit of our cost design over a baseline cost function \cite{jin2024complementarity} for avoiding local minima in complex \gls{wmm} scenarios with 100\% success rate.
Our approach also profits from no post trajectory refinement needed, no task-dependent tuning, and an easy adaptation to new robot model through geometric parametrization. 

Our work currently addresses only planar manipulation, although our representation is already partially generalizable to 3D. 
Also, it does not scale yet to real-time Model Predictive Control (MPC). 
Our future plans include adapting our representation to 3D scenarios and studying ways of speeding up our planning to tackle real-time dynamic tasks.

\section{RELATED WORKS}
\label{section:related-works}

\subsection{Contact surface representation}
\label{subsection:rl_representation}

To enable solving \gls{wmm} planning as continuous \gls{nlp}, it is necessary to have a differentiable representation of the proximity points between the robot and object surfaces.
The GJK \cite{gjk_org} and EPA \cite{van2001proximity} algorithms are commonly used in computer graphics to compute the proximity between two convex sets very quickly.
However, these algorithms have conditional steps and are not differentiable and thus cannot be implemented in a continuous \gls{nlp}.

The \gls{sdf} \cite{368173} provides the distance of a query point to a surface defined by a mathematical function. 
It has proven to be efficient to formulate collision-free motion planning for robots \cite{10611674}. 
It can be used to model contact between object and environment in contact-rich manipulation  \cite{10938330} but requires the robot kino-dynamic constraints to be abstracted away.
Combining the \gls{sdf} with a one-step gradient descent provides an estimate of the projection of a query point onto the object surface \cite{contactsdf}, but still requires sampling discrete query points on the robot surface.
Similarly, the robot body can be modeled with \gls{sdf} allowing planning contact with any part of the body, but then requires the user to pre-define the contact point on the object surface \cite{10611674}.
These pre-defined samples work well for small surfaces like robotic hands \cite{contactsdf, jin2024complementarity} but are unrealistic for covering larger surfaces like a robot body and the object at the same time.
It also requires to manually predefine the locations where the contact could happen.

In \cite{7487300} Catmull-Clark Surfaces ar used to represent the robot links together with a ray-casting algorithm to compute the proximity points to an object surface. This method offers more flexibility by optimizing contact between an object surface an a pre-selected robot part. However, the complexity of the computation of proximity points prevents its use in a continuous \gls{nlp} for \gls{wmm} planning.

In \cite{mordatch2012}, Mordatch et al. proposed a representation of robot and object bodies as patches of capsules, equivalent to rounded segments.
With the use of additional optimization variables per potential contact pair, it allows to estimate proximity points on the robot and object surfaces.
Nonetheless, this method neglects contact modes such as sliding.

Therefore, the current state of the art provides solutions that are only applicable to subsets of whole-body contacts, without generalizing to continuous optimization on the full body. In this work we address this representation problem by seeking closed-form of proximity points computation to enable formulation of continuous \gls{nlp} for \gls{wmm}.

\subsection{Contact Optimization}
\label{subsection:rl_contactoptimization}

Contact optimization is essential for improving the performance of a task \cite{LI2016352, ad27e485b48f4be48bc2a566583a7604}.
Given a continuous representation that enables integration of proximity points computation into a continuous \gls{nlp}, remains the question of how to design the cost function to encourage useful contacts for the manipulation without restricting the use of the whole-body.

In \cite{jin2024complementarity, contactsdf}, the authors rely on two cost functions to guide the manipulation planning. 
The first contact cost tries to minimize difference between the object center and sampled contact points on the robot body surface.
This cost design depends on the initial configuration of the plant and is prone to being trapped in local minima. 
Another grasp cost promotes postures that are equally distributed on the object geometry.
However, this requires to decide the number of contacts prior to contact, which is incompatible with autonomous decision in unknown situation.

Similarly, \cite{pang2023globalplanningcontactrichmanipulation} relies on Eigen grasp primitives \cite{Aburub_2025} for generating contact with the object. 
This requires to pre-compute primitives that are robot-dependent and focusing on grasps, preventing the planning of complex contact poses like using the outside surface of a single arm or a torso.

To guide the search, several works use a time horizon to plan manipulation through contacts based on its future horizon outcome \cite{moura2022nonprehensileplanarmanipulationtrajectory, 10856359, 10614849}. 
However, to remain efficient these methods consider only simple surfaces of contact at the end-effector, failing to enforce the full kino-dynamic constraints of the robot that are essential for \gls{wmm}.
An approach based on Iterative Linear Quadratic Regulator can plan manipulation trajectories using the full surface of a robot arm \cite{9981686}.
Yet, it only allows to solve local trajectories from configurations where the robot is already near contact and works only with elastic objects.
Planning \gls{wmm} trajectories with an horizon in a reasonable time is thus infeasible, and we need instead metrics that can help guiding the contact planning at lower computational cost.

Instead of using an explicit metric, learning-based methods could solve the contact selection (e.g. using affordance \cite{8239542}), yet these technics still limit to systems without kino-dynamic constraints \cite{jeon2023learningwholebodymanipulationquadrupedal}, simplified hand geometries \cite{chen2021generalinhandobjectreorientation}, or require a model-based planning input to converge \cite{zhang2023planguidedreinforcementlearningwholebody}.

Among available model-based metrics, Yoshikawa's manipulability \cite{yoshikawa1985} is often used.
In this work we propose to complete this robot-centric metric with an object-centric metric to guide the \gls{wmm}.

\section{PROBLEM FORMULATION}
\label{section:problem-formulation}

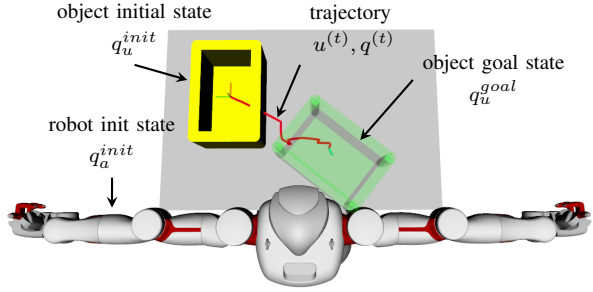
\begin{figure}[t!]
    \centering

    \begin{tikzpicture}
        \node (image) 
            {\includegraphics[width=0.9\linewidth]{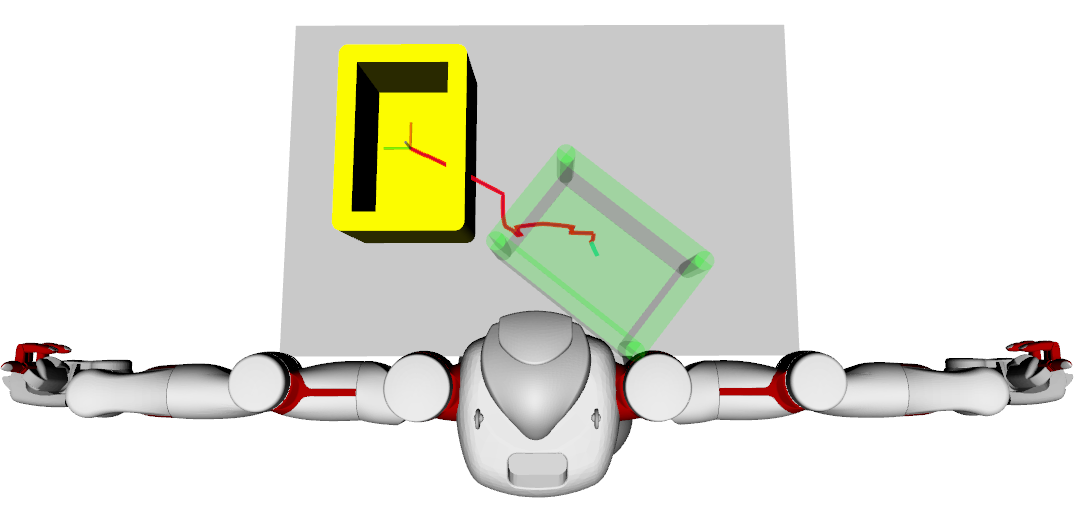}};

        \draw [arrow] (-2.2, 1.2) node [inout, anchor=south] 
            {\footnotesize{object initial state} \\ \footnotesize{$q_u^{init}$}}
            --  (-1.5, 1.0) ;

        \draw [arrow] (1.5, 1.0) node [inout, anchor=west] 
            {\footnotesize{object goal state} \\ \footnotesize{$q_u^{goal}$}}
            --  (0.8, 0.2) ;

        \draw [arrow] (-2.5, -0.3) node [inout, anchor=south] 
            {\footnotesize{robot init state} \\ \footnotesize{$q_a^{init}$}}
            --  (-2.5, -0.7) ;

        \draw [arrow] (0.0, 1.2) node [inout, anchor=south west] 
            {\footnotesize{trajectory } \\ \footnotesize{$u^{(t)}, q^{(t)}$}}
            --  (-0.3, 0.5) ;
            
    \end{tikzpicture}
    \caption{Overview of the \gls{wmm} planning problem.}
    \label{fig:planning_drawing}
\end{figure}

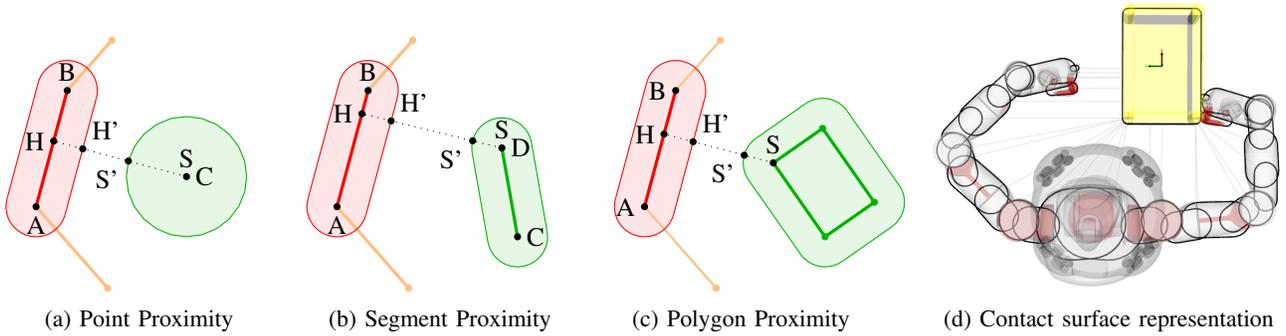
\begin{figure*}[t]

    \def\objcol{black!30!green}
    \def\bckcol{orange!50}
    \def\robcol{red}
    
    \centering
    \hfill
    \begin{minipage}[b]{0.2\textwidth}
        \vspace*{0mm}
        \begin{tikzpicture}[scale=0.4]
        \begin{scope}[shift={(-3,2)}, rotate=-15]
            \fill[\robcol!10,domain=180:360] plot ({cos(\x)}, {sin(\x)});
            \fill[\robcol!10,domain=0:180] plot ({cos(\x)}, {4+sin(\x)});
            \fill[\robcol!10] (-1, 0) rectangle (1, 4);
        
            \draw[\bckcol,very thick] (3,-2) -- (0, 0);
            \draw[\robcol,very thick] (0,0) -- (0, 4);
            \draw[\bckcol,very thick] (0,4) -- (1, 6);
            
            \draw[\bckcol,fill=\bckcol] (3,-2) circle (0.1cm);
            \draw[black,fill=black] (0,0) circle (0.1cm) node[anchor=north] {A};
            \draw[black,fill=black] (0,4) circle (0.1cm) node[anchor=south] {B};
            \draw[\bckcol,fill=\bckcol] (1,6) circle (0.1cm);

            \draw[\robcol,domain=180:360] plot ({cos(\x)}, {sin(\x)});
            \draw[\robcol,domain=0:180] plot ({cos(\x)}, {4+sin(\x)});
            \draw[\robcol] (1, 0) -- (1, 4);
            \draw[\robcol] (-1, 0) -- (-1, 4);
        \end{scope}

        \begin{scope}[shift={(2,3)}]
            \fill[\objcol!10,domain=0:360,thick] plot ({2*cos(\x)}, {2*sin(\x)});

            \draw[black,fill=black] (0,0) circle (0.1cm) 
                node[anchor=west] {C} node[anchor=south] {S};

            \draw[\objcol,domain=0:360] plot ({2*cos(\x)}, {2*sin(\x)});
        \end{scope}

        \draw[black,dotted] (2, 3) -- (-2.41506, 4.18301);
        \draw[black,fill=black] (-2.41506, 4.18301) circle (0.1cm) 
            node[anchor=east] {H};
        \draw[black,fill=black] (-1.44914, 3.92419) circle (0.1cm) 
            node[anchor=south west] {H'};
        \draw[black,fill=black] (0.0681483, 3.51764) circle (0.1cm) 
            node[anchor=north east] {S'};
        
        \end{tikzpicture}

        \centering
        \subcaption{Point Proximity}
        \label{fig:representation_point}
    \end{minipage}
    \hfill
    \begin{minipage}[b]{0.2\textwidth}
        \vspace*{0mm}
        \begin{tikzpicture}[scale=0.4]
        \begin{scope}[shift={(-3,2)}, rotate=-15]
            \fill[\robcol!10,domain=180:360] plot ({cos(\x)}, {sin(\x)});
            \fill[\robcol!10,domain=0:180] plot ({cos(\x)}, {4+sin(\x)});
            \fill[\robcol!10] (-1, 0) rectangle (1, 4);
        
            \draw[\bckcol,very thick] (3,-2) -- (0, 0);
            \draw[\robcol,very thick] (0,0) -- (0, 4);
            \draw[\bckcol,very thick] (0,4) -- (1, 6);
            
            \draw[\bckcol,fill=\bckcol] (3,-2) circle (0.1cm);
            \draw[black,fill=black] (0,0) circle (0.1cm) node[anchor=north] {A};
            \draw[black,fill=black] (0,4) circle (0.1cm) node[anchor=south] {B};
            \draw[\bckcol,fill=\bckcol] (1,6) circle (0.1cm);

            \draw[\robcol,domain=180:360] plot ({cos(\x)}, {sin(\x)});
            \draw[\robcol,domain=0:180] plot ({cos(\x)}, {4+sin(\x)});
            \draw[\robcol] (1, 0) -- (1, 4);
            \draw[\robcol] (-1, 0) -- (-1, 4);
        \end{scope}

        \begin{scope}[shift={(3,1)}, rotate=10]
            \fill[\objcol!10,domain=180:360,thick] plot ({cos(\x)}, {sin(\x)});
            \fill[\objcol!10,domain=0:180,thick] plot ({cos(\x)}, {3+sin(\x)});
            \fill[\objcol!10,thick] (-1, 0) rectangle (1, 3);
        
            \draw[\objcol,very thick] (0,0) -- (0, 3);
            \draw[black,fill=black] (0,0) circle (0.1cm) 
                node[anchor=west] {C};
            \draw[black,fill=black] (0,3) circle (0.1cm) 
                node[anchor=west] {D} node[anchor=south] {S};

            \draw[\objcol,domain=180:360] plot ({cos(\x)}, {sin(\x)});
            \draw[\objcol,domain=0:180] plot ({cos(\x)}, {3+sin(\x)});
            \draw[\objcol] (-1, 0) -- (-1, 3);
            \draw[\objcol] (1, 0) -- (1, 3);
        \end{scope}

        \draw[black,dotted] (2.47906, 3.95442) -- (-2.17178, 5.09096);
        \draw[black,fill=black] (-2.17178, 5.09096) circle (0.1cm) 
            node[anchor=east] {H};
        \draw[black,fill=black] (-1.20037, 4.85357) circle (0.1cm) 
            node[anchor=south west] {H'};
        \draw[black,fill=black] (1.50765, 4.19181) circle (0.1cm) 
            node[anchor=north east] {S'};
        
        \end{tikzpicture}

        \centering
        \subcaption{Segment Proximity}
        \label{fig:representation_segment}
    \end{minipage}
    \hfill
    \begin{minipage}[b]{0.2\textwidth}
        \vspace*{0mm}
        \begin{tikzpicture}[scale=0.4]
        \begin{scope}[shift={(-3,2)}, rotate=-15]
            \fill[\robcol!10,domain=180:360] plot ({cos(\x)}, {sin(\x)});
            \fill[\robcol!10,domain=0:180] plot ({cos(\x)}, {4+sin(\x)});
            \fill[\robcol!10] (-1, 0) rectangle (1, 4);
        
            \draw[\bckcol,thick] (3,-2) -- (0, 0);
            \draw[\robcol,very thick] (0,0) -- (0, 4);
            \draw[\bckcol,thick] (0,4) -- (1, 6);
            
            \draw[\bckcol,fill=\bckcol] (3,-2) circle (0.1cm);
            \draw[black,fill=black] (0,0) circle (0.1cm) node[anchor=east] {A};
            \draw[black,fill=black] (0,4) circle (0.1cm) node[anchor=east] {B};
            \draw[\bckcol,fill=\bckcol] (1,6) circle (0.1cm);

            \draw[\robcol,domain=180:360] plot ({cos(\x)}, {sin(\x)});
            \draw[\robcol,domain=0:180] plot ({cos(\x)}, {4+sin(\x)});
            \draw[\robcol] (1, 0) -- (1, 4);
            \draw[\robcol] (-1, 0) -- (-1, 4);
        \end{scope}

        \begin{scope}[shift={(3,1)}, rotate=35]
            \fill[\objcol!10,domain=180:270] (0, 0) -- plot ({cos(\x)}, {sin(\x)});
            \fill[\objcol!10,domain=270:360] (2, 0) -- plot ({2+cos(\x)}, {sin(\x)});
            \fill[\objcol!10,domain=0:90] (2, 3) -- plot ({2+cos(\x)}, {3+sin(\x)});
            \fill[\objcol!10,domain=90:180] (0, 3) -- plot ({cos(\x)}, {3+sin(\x)});
            \fill[\objcol!10] (0, 0) rectangle (2, 3);
            \fill[\objcol!10] (2, 0) rectangle (3, 3);
            \fill[\objcol!10] (2, 3) rectangle (0, 4);
            \fill[\objcol!10] (-1, 3) rectangle (0, 0);
            \fill[\objcol!10] (0, -1) rectangle (2, 0);
        
            \draw[\objcol,very thick] (0,0) rectangle (2, 3);
            \draw[\objcol,fill=\objcol] (0,0) circle (0.1cm);
            \draw[black,fill=black] (0,3) circle (0.1cm) node[anchor=south] {S};
            \draw[\objcol,fill=\objcol] (2,0) circle (0.1cm);
            \draw[\objcol,fill=\objcol] (2,3) circle (0.1cm);

            \draw[\objcol,domain=180:270] plot ({cos(\x)}, {sin(\x)});
            \draw[\objcol,domain=270:360] plot ({2+cos(\x)}, {sin(\x)});
            \draw[\objcol,domain=0:90] plot ({2+cos(\x)}, {3+sin(\x)});
            \draw[\objcol,domain=90:180] plot ({cos(\x)}, {3+sin(\x)});
            \draw[\objcol] (3, 0) -- (3, 3);
            \draw[\objcol] (2, 4) -- (0, 4);
            \draw[\objcol] (-1, 3) -- (-1, 0);
            \draw[\objcol] (0, -1) -- (2, -1);
        \end{scope}

        \draw[black,dotted] (1.27927, 3.45746) -- (-2.35295, 4.41481);
        \draw[black,fill=black] (-2.35295, 4.41481) circle (0.1cm) node[anchor=east] {H};
        \draw[black,fill=black] (-1.38597, 4.15994) circle (0.1cm) node[anchor=south west] {H'};
        \draw[black,fill=black] (0.312294, 3.71233) circle (0.1cm) node[anchor=north east] {S'};
        
        \end{tikzpicture}

        \centering
        \subcaption{Polygon Proximity}
        \label{fig:representation_polygon}
    \end{minipage}
    \hfill
    \begin{minipage}[b]{0.3\textwidth}
        \vspace*{0mm}
        \centering
        \includegraphics[width=\textwidth,trim={1cm 0 1cm 0},clip]
            {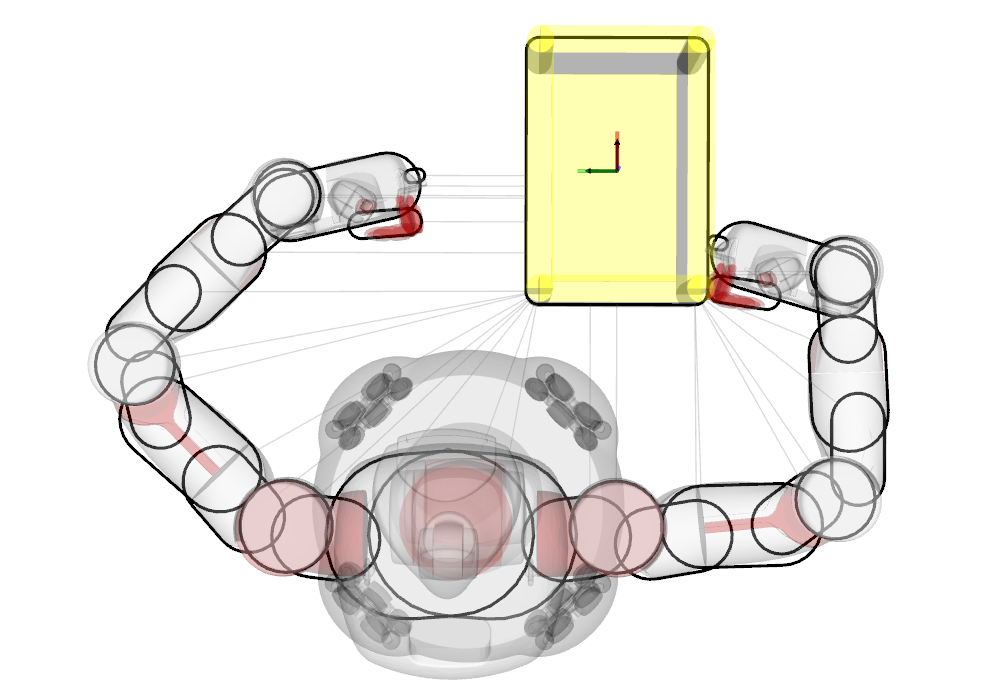}
        \subcaption{Contact surface representation}
        \label{fig:representation_patches}
    \end{minipage}
    \hfill

    \centering
    \captionsetup{justification=justified,margin=0.0cm}
    \caption{Representation of the contact surface for the robot and object. The robot contact surface is represented by a set of rounded segment patches (d). The object can be represented by a circle (a), a rounded segment (b) or a rounded polygon (c)(d). The collision avoidance between robot parts pairs is represented by two rounded segments (b).}
    \label{fig:representation}
    
\end{figure*}

The \gls{wmm} planning problem we address in this work consists in finding a trajectory of robot actions $u^{(t)}$ that brings an object from an initial state $q_u^{init}$ to a goal state $q_u^{goal}$, the robot starting in the state $q_a^{init}$ (see \cref{fig:planning_drawing}).
We assume the manipulation to be planar, the dynamics to be quasi-dynamics and the object to be convex.
We also assume the planning to take into account all the kino-dynamic constrains of the system, namely the robot joint and torque limits, the robot self-collision avoidance, the object workspace boundaries and the dynamics of the different contact modes.

\subsection{State and Input}
\label{subsection:state-tinput}

We define the state of the plant $q$ as the vertical concatenation of the object state $q_u \in {\rm I\!R}^{3 \times 1}$ and the robot state $q_a \in {\rm I\!R}^{N_a \times 1}$, where $N_a$ is the number of active joints of the robot. 
Working in the plane, the state of the objects comprises two position variables in the $xy$-plane, and one rotation variable along the $z$-axis.
The control input vector of the robot is denoted $u \in {\rm I\!R}^{N_a \times 1}$ and corresponds to the displacement command of the robot in impedance control.
The robot is constrained to move in the same $xy$-plane as the object and only joints rotating along the $z$-axis are active and part of the planning.

\subsection{Quasi-Dynamic State Transition}
\label{subsection:state-transition}

The state transition computes the next plant state $q^+$ based on the current state $q$ and the control input $u$.
Similar to \cite{pang2023globalplanningcontactrichmanipulation}, we assume a quasi-dynamic model for the system, which ignores the acceleration terms, and consider the a limit surface model for the object dynamics \cite{hogan2020reactive}, resulting in the following linear equation for system's state transition
\begin{subequations}
\begin{align}
    \mathbf{M} (q^+ - q) &= \mathbf{K} u + J(q)^{\top} f \\
    \text{s.t.} \quad
        \mathbf{M} &= \begin{pmatrix} L_u & 0_{3 \times N_a} \\ 0_{N_a \times 3} & K_a \end{pmatrix} \\
        \mathbf{K} &= \begin{pmatrix} 0_{3 \times 1} \\ K_a  \end{pmatrix} \\
        J^{\top} f &= \sum_{k \in \mathcal{C}} \begin{pmatrix} J_{u,k}^{\top} \\ J_{a,k}^{\top} \end{pmatrix} f_k,
\end{align}
\label{eq:quasi-dynamics}%
\end{subequations}
where $L_u \in {\rm I\!R}^{3 \times 3}$ corresponds to the friction matrix between the object and the table, $K_a \in {\rm I\!R}^{N_a \times N_a}$ the impedance gain matrix of the robot joints, $\mathcal{C}$ to the set of active contact points, $f_k \in {\rm I\!R}^{2 \times 1}$ the contact forces and $J_{u,k} \in {\rm I\!R}^{3 \times 2}$ and $J_{a,k} \in {\rm I\!R}^{N_a \times 2}$ the Jacobians of the $k^{th}$ contact location respectively in the object frame and through the robot kinematic chain.

Each contact force $f_k=(f_k^n; f_k^t)$, where $f_k^n$ is the normal component of the force and $f_k^n$ its tangential components, must satisfy the complementarity constraints and cone constraints of the Coulomb friction
\begin{subequations}
\begin{align}
        -\mu f_k^n &\leq f_k^t \leq \mu f_k^n \\
        f_k^n &\geq 0 \\
        v_k^n f_k^n &= 0 \\
        v_k^t (|f_k^t| - \mu f_k^n) &= 0 \\
        v_k^t f_k^t &\geq 0,
\end{align}
\label{eq:force-coulomb}%
\end{subequations}
with $\mu$ the Coulomb friction coefficient - assumed uniform on all surfaces - and $v_k^n$ and $v_k^t$ respectively the normal and tangential component of the contact point velocity $v_k$.

\subsection{Surface Proximity}
\label{subsection:surface-proximity}

To enable integrating the contact location on the robot and object surfaces in continuous optimization, we need to compute the proximity points between these surfaces in a differentiable manner.
The proximity problem consists in finding the two closest points H on a surface $\mathcal{S}_H$ and S on a surface $\mathcal{S}_S$, whose distance d corresponds to the minimum distance between the two surfaces:
\begin{equation}
  \begin{aligned}
    (H, S) &= \argmin_{P_1 \in \mathcal{S}_H, P_2 \in \mathcal{S}_S}{d(P_1, P_2)} \\
    \text{with} \quad d &= || P_1 P_2 ||.
    \label{eq:proximity}%
  \end{aligned}
\end{equation}
Taking into account how the surfaces $\mathcal{S}_H$ and $\mathcal{S}_S$ move with the object and robot states $q$, we can formulate a proximity function $(H, S)=\psi(\mathcal{S}_H, \mathcal{S}_S, q)$. To permit continuous optimization on contact location, we need this function to be continuous and differentiable.

\section{METHOD}
\label{section:method}

\subsection{Robot and object surface representations}

We present here our method to compute the proximity points between a robot surface and the object surface, which novelty is to be formulated as a differentiable closed-form.
Our formulation relies on three key elements: a representation of robot and object geometries as patches of structures (point, segment or convex polygon); a differentiable closed-form solution to compute the proximity points between the different structures; a rounding that transforms the structures into rounded surfaces and provides the proximity points on these surfaces.

The robot is represented by a set of rounded segment patches as shown in \cref{fig:representation}. 
Each patch is attached to a robot link and moves according to the joints configuration $q_a$.

We use two different sets of patches to represent the robot geomtry: one set $\mathcal{B}_r$ with several patches per link to represent the surface that may contact with the object - which needs to accurately fit the robot shape; and one set $\mathcal{B}_c$ with one patch per link covering the full link used for self-collision avoidance - which can be more conservative.

The structure of each patch used to represent the robot is a segment. Noting A and B the two ends of the segment, we can define a point H on this segment with a parameter h as  
\begin{equation}
  \begin{aligned}
    &AH = h AB \\
    &\text{with} \quad 0 \leq h \leq 1.
    \label{eq:segmentAB}%
  \end{aligned}
\end{equation}

The object structure is represented either as a rounded point (\cref{fig:representation_point}), segment (\cref{fig:representation_segment}) or convex-polygon (\cref{fig:representation_polygon}).
We detail below our computation of proximity points for different pairs of structure shapes.

\subsubsection{Point Proximity (Circle)}

The computation of proximity points between a segment AB - representing the robot - and an external point C - representing the object - (\cref{fig:representation_point}) is straightforward :
\begin{equation}
  \begin{aligned}
    h &= \biggr\langle \frac{AB^\top AC}
                            {AB^\top AB} \biggr\rangle_0^1,
    \label{eq:proximity_point}%
  \end{aligned}
\end{equation}
where we define the operator $\langle x \rangle_a^b = \min{(b, \max{(a, x)})}$. 
The projection H on AB is obtained from \cref{eq:segmentAB} and the projection S is the point C itself.
This representation is useful for the manipulation of an object with circular shape.

\subsubsection{Rounding}

After computing the proximity points H and S, we can round the structures by moving the projections H and S towards each other by the rounding values $r_h$ for H and $r_s$ for S.
This provides the final proximity points H' on the robot surface and S' on the object surface:
\begin{subequations}
\begin{align}
    HH' &= r_h \frac{HS}{||HS||} \\
    SS' &= r_s \frac{SH}{||SH||} \\
    d' &= d - r_h - r_s.
\end{align}
\label{eq:rounding}%
\end{subequations}

\subsubsection{Segment Proximity}
\label{subsubsection:segment_proximity}

The segment proximity algorithm computes the proximity points $H \in AB$ and $S \in CD$ between the two segments AB and CD. 
The projection S on CD is defined as
\begin{equation}
  \begin{aligned}
    CS &= s CD \\
    \text{with} \quad & 0 \leq s \leq 1.
    \label{eq:segmentCD}%
  \end{aligned}
\end{equation}
By computing the following three steps in \cref{eq:sdf-seg-seg}, we can always find a unique solution for proximity points H and S, even if the segments are intersecting or parallel. 
The small positive regularization term $\varepsilon$ (typically $1e^{-5}m^2$) purpose is to reduce the infinity of solutions when the two segments are parallel to a unique solution, while covering the cases where a segment length could be null. 
The three steps of the algorithm consist in computing the parameters $h_0$, $s$ and $h$ following the \cref{eq:sdf-seg-seg}. 
The proximity points H and S are retrieved from parameters $h$ and $s$ with \cref{eq:segmentAB} and \cref{eq:segmentCD}.
\begin{subequations}
\begin{align}
    h_0 &= \biggr\langle \frac{AB^\top [CD]^2 AC}
                              {AB^\top [CD]^2 AB + \varepsilon} \biggr\rangle_0^1  \\
    s &= \biggr\langle \frac{CD^\top (h_0 AB - AC)}
                            {CD^\top CD + \varepsilon} \biggr\rangle_0^1 \\
    h &= \biggr\langle \frac{AB^\top (s CD + AC)}
                            {AB^\top AB + \varepsilon} \biggr\rangle_0^1
\end{align}
\label{eq:sdf-seg-seg}%
\end{subequations}

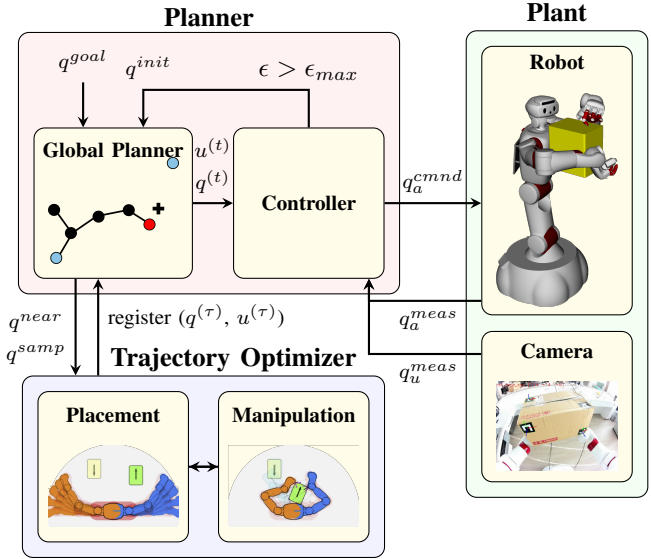
\begin{figure}[ht]
    \centering
    \hfill
    \begin{tikzpicture}[node distance=3.3cm]
        \node (tree) [bloc, fill=yellow!10] {\textbf{Global Planner} \\ 
            \includegraphics[width=0.2\linewidth]{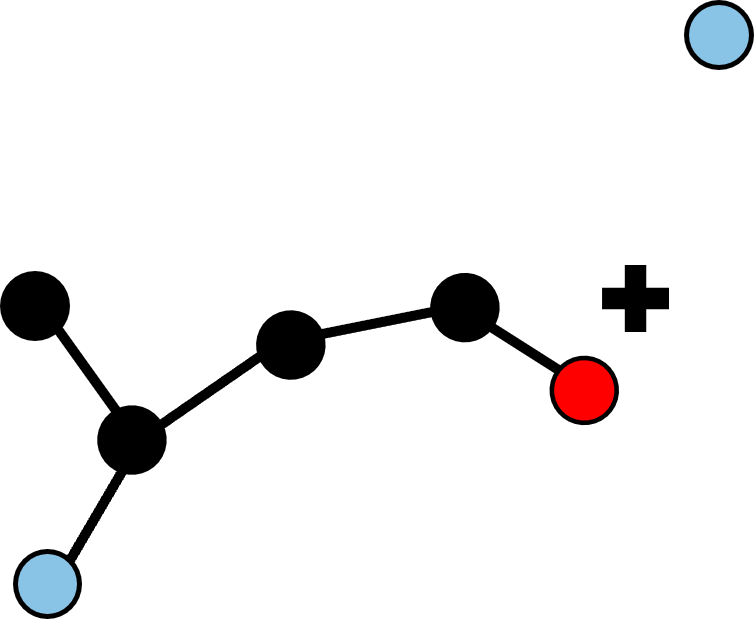}};
        
        \node (cfree) [bloc, draw=black!100, minimum height=2.0cm] at ([shift={(0,-2.5)}]tree) {\textbf{Placement} \\\\
            \includegraphics[width=0.2\linewidth]{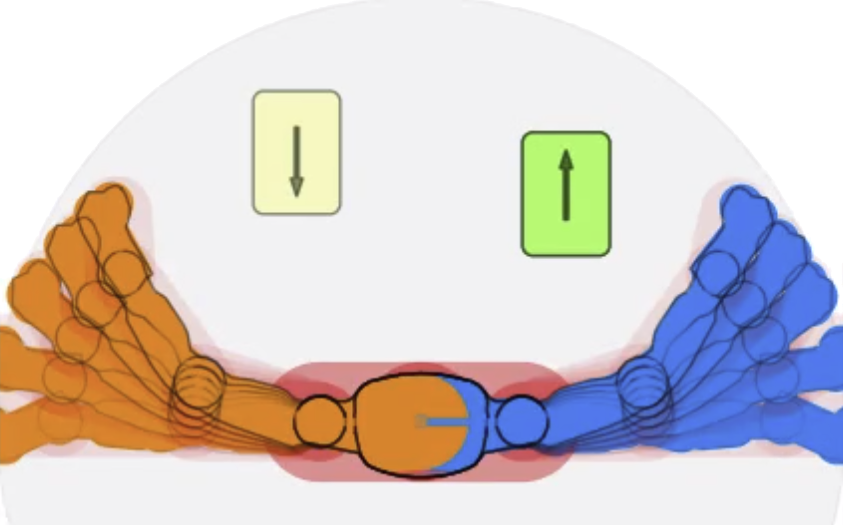}};

        \node (crich) [bloc, draw=black!100, minimum height=2.0cm] at ([shift={(2.4,-2.5)}]tree) {\textbf{Manipulation} \\\\
            \includegraphics[width=0.2\linewidth]{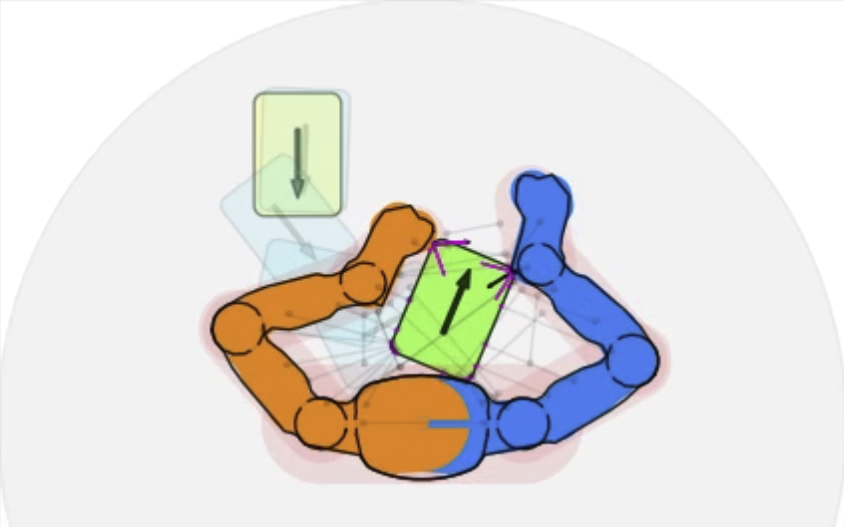}};

        \node (control) [bloc, right of=tree, node distance=2.6cm] {\textbf{Controller}};
        
        \node (robot) [bloc, right of=control, shift={(0.0, 0.3)}] {\textbf{Robot} \\\\
            \includegraphics[width=.09\textwidth]{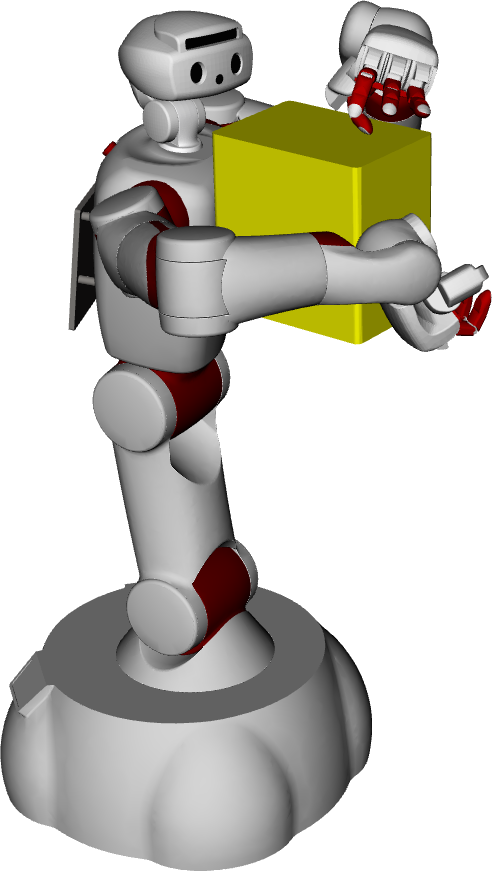} };
            
        \node (camera) [bloc] at ([shift={(0.0,-2)}]robot) 
            {\textbf{Camera} \\\\ \includegraphics[width=.09\textwidth]{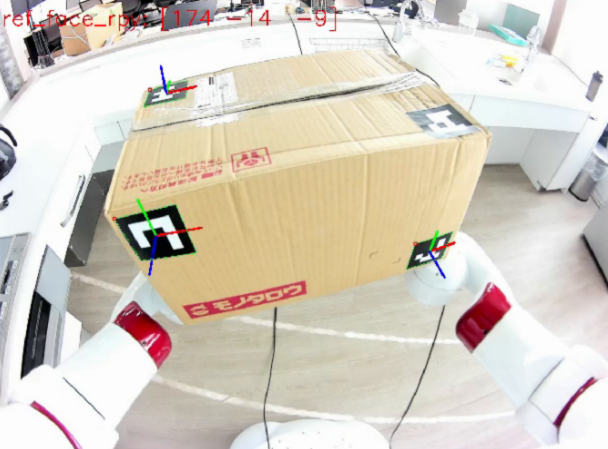} 
            };

        \node (readtree) [inout, above of=tree, font=\footnotesize, anchor=east, shift={(0.9, 0.3)}] {\footnotesize{$q^{init}$}};

        \begin{scope}[on background layer]
            \node[fit=(cfree) (crich), draw, inner sep=0.2cm, rounded corners, fill=blue!5] (trajopt) { };
            \node (trajopttitle) [inout, above of=trajopt, anchor=north, yshift=+0.2cm, xshift=0.4cm] {\textbf{Trajectory Optimizer}};
        \end{scope}

        \begin{scope}[on background layer]
            \node[fit=(robot) (camera), draw, inner sep=0.2cm, rounded corners, fill=green!5] (plant) { };
            \node (planttitle) [inout, above of=plant, anchor=south, yshift=1.6cm] {\textbf{Plant}};
        \end{scope}

        \begin{scope}[on background layer]
            \node[fit=(tree) (control) (readtree), draw, inner sep=0.2cm, rounded corners, fill=red!5] (planner) { };
            \node (plannertitle) [inout, above of=planner, anchor=south, yshift=0.2cm] {\textbf{Planner}};
        \end{scope}

        \draw [arrow] (tree.north) ++(-0.4, 0.6) -- 
            node [inout, anchor=south, shift={(0.0, 0.3)}] 
            {\footnotesize{$q^{goal}$}} 
            ([shift={(-0.4, 0.0)}]tree.north);
        
        \draw [arrow] ([shift={(-0.5,0.0)}]tree.south) 
            node [inout, anchor=east] at ([shift={(-0.5,-0.8)}]tree.south)
            {\footnotesize{$q^{near}$} \\ \footnotesize{$q^{samp}$}} 
            -- ([shift={(-1.7,0.0)}]trajopt.north);
        \draw [arrow] ([shift={(-1.4, 0.0)}]trajopt.north) 
            node [inout, anchor=west] at ([shift={(-0.2, -0.5)}]tree.south)
            {\footnotesize{register ($q^{(\tau)}$, $u^{(\tau)}$)}} 
            -- ([shift={(-0.2, 0.0)}]tree.south);
        
        \draw [arrow] (tree.east) -- node [inout, anchor=south] 
            {\footnotesize{$u^{(t)}$} \\ \footnotesize{$q^{(t)}$}} (control.west) ;
        \draw [arrow] (control.east) -- node [inout, anchor=south] 
            {\footnotesize{$q_a^{cmnd}$}} ([shift={(0.0, -0.3)}]robot.west) ;
        \draw [arrow] (control.north) -| ++(0.0, 0.5) node [inout, anchor=south] {$\epsilon > \epsilon_{max}$} -| ([shift={(0.4, 0.0)}]tree.north) ;
        \draw [arrow] ([shift={(0.0, 0.7)}]camera.west) -- ++(-1.2, 0.0) 
            node [inout, anchor=north, shift={(0.5, 0.0)}] 
            {\footnotesize{$q_u^{meas}$}} 
            -| ([shift={(0.8, 0.0)}]control.south) ;
        \draw [arrow] ([shift={(0.0, -0.1)}]robot.west) ++(0.0, -1.5) -- ++(-1.5, 0.0) 
            node [inout, anchor=north, shift={(0.8, 0.0)}] 
            {\footnotesize{$q_a^{meas}$}} 
            -| ([shift={(0.8, 0.0)}]control.south) ;
        
        \draw[arrow] (cfree.east) -- (crich.west);
        \draw[arrow] (crich.west) -- (cfree.east);
        
    \end{tikzpicture}
    \hfill
    \centering
    \captionsetup{justification=centering, margin=1.0cm}
    \caption{Pipeline of the proposed planner}
    \label{pipeline}
\end{figure}

\subsubsection{Polygon Proximity}

We propose an algorithm with a fixed number of step to find the closest points between a segment and a convex polygon $\mathcal{P}=(P_1, ...,P_{N_p})$. 
For a convex polygon made of $N_p$ edges, this algorithm allows to reduce the number of contact parameters $(h,s)$ from $N_p$ to 1.
The algorithm consists of $N_p-1$ steps \cref{alg:polygon-proximity} that relies on the segment proximity process $segmentproximity$ (\cref{subsubsection:segment_proximity}) to reduce the number of edges of the polygon by one at each step, until it simplifies to a single segment to segment proximity problem.

\algrenewcommand\algorithmicrequire{\textbf{Input:}}
\algrenewcommand\algorithmicensure{\textbf{Output:}}

\begin{algorithm}[h!]
\begin{algorithmic}
    \Require $A, B, \mathcal{P}=(P_1, ...,P_{N_p})$
    \Ensure $(H, S) = \argmin_{P_h \in AB, P_s \in \mathcal{P}}{||P_h P_s||}$
    \State $\mathcal{P}_{new} \gets (P_1, ...,P_{N_p})$
    \For{$i \gets 1$ to $N_p-1$}
        \State $\mathcal{P}_{tmp} \gets \mathcal{P}_{new}$
        \For{$j \gets 1$ to $N_p-1-i$}
            \State $(C, D) \gets (P_{tmp}[j], P_{tmp}[j+1])$
            \State $(P_h,P_s) \gets segmentproximity(A, B, C, D)$
            \State Append $S$ to $\mathcal{P}_{new}$
        \EndFor
    \EndFor
    \State $(H,S) \gets (P_h,P_s)$.
\end{algorithmic}
\caption{Polygon Proximity Algorithm}\label{alg:polygon-proximity}
\end{algorithm}

\subsection{Pipeline}

\begin{figure*}[th]
    \centering
    \begin{tabular}{ |l r||r r r r r|r r r r r| }
         \hline
            \multicolumn{2}{|c|}{\textbf{Planning time [mn]}}
            & \multicolumn{5}{|c|}{Scenario 1} 
            & \multicolumn{5}{|c|}{Scenario 2} \\ 
         \hline
         \multicolumn{2}{|l|}{$q_u^{init}$ [m, m, deg]}
            & \multicolumn{5}{|c|}{(0.60, -0.20, 0.00)} 
            & \multicolumn{5}{|c|}{(0.65,  0.00, 0.00)} \\ 
         \hline
         \multicolumn{2}{|l|}{$q_a^{init}$ [deg]}
            & \multicolumn{5}{|c|}{(-40.1, 0.0, 0.0, 40.1, 0.0, 0.00)} 
            & \multicolumn{5}{|c|}{(0.0, 0.0, 0.0, 0.0, 0.0, 0.0)} \\ 
         \hline
         \multicolumn{2}{|l|}{$q_u^{goal}$ [m, m, deg]}
            & \multicolumn{5}{|c|}{(0.60, 0.20, $\theta$)} 
            & \multicolumn{5}{|c|}{(0.35, 0.00, $\theta$)} \\ 
         \hline
         \multicolumn{2}{|l|}{$\theta$ [deg]}
            & 0 & 45 & 90 & 135 & 180 
            & 0 & 45 & 90 & 135 & 180 \\
         \hline
         \hline
         OURS 
            & $\text{success}^1$ & 100\% & 100\% & 100\% & 100\% & 93\%
                      & 100\% & 100\% & 100\% & 100\% & 100\% \\
            & mean [mn]& 
                \textbf{1.2} & 
                \textbf{1.9} & 
                \textbf{5.7} & 
                \textbf{6.5} & 
                \textbf{16.0} & 
                \textbf{0.5} & 
                \textbf{0.4} & 
                \textbf{5.3} &
                \textbf{7.8} & 
                \textbf{5.3}  \\
            & min [mn]
                & 1.0 & 1.7 & 3.4 & 4.6 & 6.0 
                & 0.3 & 0.4 & 4.6 & 4.5 & 3.1 \\
            & max [mn]
                & 1.3 & 2.1 & 11.2 & 10.1 & - 
                & 0.7 & 0.5 & 8.1 & 12.5 & 12.9 \\
        \hline
        Baseline 
            & $\text{success}^1$ 
                & 100\% & 70\% & 7\% & 0\% & 0\%
                & 93\% & 40\% & 33\% & 7\% & 0\% \\
            & mean [mn]
                & 3.9 & 13.5 & 29.1 & - & - 
                & 9.7 & 20.9 & 22.7 & 28.7 & - \\
            & min [mn]
                & 0.3 & 0.6 & 5.9 & - & -
                & 0.4 & 0.3 & 0.5 & 2.7 & - \\
            & max [mn]
                & 26.1 & - & - & - & - 
                & - & - & - & - & - \\
         \hline
    \end{tabular}
    
    \centering
    \captionsetup{justification=justified,margin=0.0cm}
    \caption{Comparison of planning performance our method (OURS) and the baseline sampling planner \cite{pang2023globalplanningcontactrichmanipulation} (Baseline). \\
    $^{(1)}$ Planning time exceeding the timeout of 30mn is considered as failure.}
    \label{fig:experiment_comparison}
\end{figure*}

\subsubsection{Global Planner}

The role of the global planner is to ensure that the planning does not get stuck in local minimum and find a solution even for complicated scenarios. 
It is implemented as a \gls{rrt} with rewiring of similar nodes, building a graph of explored system states by iterating through the following steps: 

\textbf{Sample}: samples a sub-goal state $q^{samp}$ for the system based on a greed parameter between 0 and 1: if the greed is 1 the planner always uses the final goal state $q^{goal}$ as sub-goal; if it is 0, the planner always uses a random state within the workspace; in between it chooses randomly between the two alternatives proportionally to the greed value.

\textbf{Nearest}: chooses the nearest state $q^{near}$ in the tree that is the closest to the sampled sub-goal, based on Euclidean metric or random choice.

\textbf{Extend}: extends the tree starting from the nearest state towards the sub-goal state with the trajectory result ($q^{(\tau)}$, $u^{(\tau)}$) from the trajectory optimizer.

\textbf{Register}: registers the trajectory of explored states to the tree and re-wires close nodes to avoid populating the tree with too many similar nodes. If the latest node registered is close enough to the final goal $q^{goal}$, the shortest trajectory ($q^{(t)}$, $u^{(t)}$) from the initial node $q^{init}$ to the goal is extracted using the Dijkstra’s algorithm \cite{Dijkstra1959}.

\subsubsection{Trajectory Optimizer}
\label{subsubsection:trajectory_optimizer}

The trajectory optimizer receives from the global planner the nearest node $q^{near}$ and the sub-goal $q^{samp}$. It tries to generate a trajectory that goes as close as possible to $q^{samp}$ from $q^{near}$ using sequential \gls{to}. 
For that, it first starts with a placement phase, which is a $N_{cf}$ steps contact-free \gls{to} that seeks contact with the object without applying any force. 
Once a contact is reached, the trajectory optimizer switches to a manipulation phase, a $N_{cr}$ step contact-rich \gls{to} that first detects active contacts that may apply forces to the object and optimized the robot action to manipulate the object, while allowing the making and breaking of contacts at each step.

\textbf{Placement}: The placement optimization is a \gls{to} that seeks a robot configuration which is contacting with the object while optimizing that contact location to have good manipulability of both the robot and the object. 
The optimization objective formulates as follows and applies only to the last state $q^{N_{cf}}$ of the trajectory.
\begin{subequations}
\begin{align}
    \min_{u} \quad & \mathcal{G}_p(q^{N_{cf}}) + \mathcal{G}_r(q^{N_{cf}}) \\
    \text{with} \quad 
        q^+ &= q + \mathbf{M}^{-1} \mathbf{K} u \\
        \mathcal{G}_p(q) &= -\beta_p \sum_{k \in \mathcal{C}}{w_k^p(q) w_k^d(q)} \\
        \mathcal{G}_r(q) &= -\beta_r \sum_{k \in \mathcal{C}}{w_k^r(q) w_k^d(q)} \\
        w_k^d(q) &= \ln{(1 + \exp{(-\beta_d d_k^r(q))}} \\
        w_k^p(q) &= (1 - (\frac{\phi_k(q)}{\pi})^2)
            \biggr\langle \frac{\tilde{f}_k^n(q)}{f_{lim}} \biggr\rangle_0^1 \\
        w_k^r(q)  &= \det{(J_k^r(q) {J_k^r(q)}^{\top} + \beta_j I)} \\
        \phi_k(q) &= \arctan(\tilde{f}_k^t(q), \tilde{f}_k^n(q)) \\
        \tilde{f}_k^n(q)  &= J_{u,k}^n(q) (q_u - q_u^{goal}) \\
        \tilde{f}_k^t(q)  &= J_{u,k}^t(q) (q_u - q_u^{goal}),
\end{align}
\label{eq:contact_transition}
\end{subequations}
where $d_k^r$ is the proximity distance between the object and the $k^{th}$ patch of $\mathcal{B}_r$ representing the robot surface. 
$\beta_d$, $\beta_p$, $\beta_r$ and $\beta_j$ are tunable scalar gains.
$\tilde{f}_k^n$ and $\tilde{f}_k^t$ are the normal and tangential components of a reference force $\tilde{f}_k$ to apply at the $k^{th}$ contact if it was active.
The costs $\mathcal{G}_p$ and $\mathcal{G}_r$ correspond respectively to whole-body object-centric and robot-centric manipulability metrics.
$w_k^p$ is an object-centric manipulability metric evaluated at the $k^{th}$ contact and based on the angle of reference force $\tilde{f}_k$ to the contact normal.
$w_k^r$ is a Yoshikawa's manipulability metric \cite{yoshikawa1985} evaluated at the $k^{th}$ contact and adapted to provide a strictly positive value for contacts with a rank-deficient Jacobian matrix.
$w_k^d$ is an activation function weighting how close the $k^{th}$ contact is to be active based on the distance $d_k^r$.

Also, each step of the placement enforces the constraints
\begin{subequations}
\begin{align}
    0 &\leq d_k^r \ \forall \ k \in [1 ... |\mathcal{B}_r|] \\
    0 &\leq d_k^c \ \forall \ k \in [1 ... |\mathcal{B}_c|] \\
    -u_{max} &\leq u \leq u_{max} \\
    q^{lb} &\leq q \leq q^{ub},
\end{align}
\label{eq:constraint_place}
\end{subequations}
where $d_k^c$ the distance between each pair of patch within $\mathcal{B}_c$ for self-collision avoidance.
$q_a^{lb}$ and $q_a^{ub}$ are the lower and upper bounds for the robot joints $q_a$.
$u_{max}$ is the vector of acceptable maximum displacement in the impedance controller for each joint during one state transition.

\textbf{Manipulation}: To calculate the state update $q^+$ in Manipulation phase, we introduce an optimization variable $\nu \in {\rm I\!R}^{(3+N_a) \times N_{cr}}$ that corresponds to a virtual motion of the system. 
Based on quasi-dynamics equations \cref{eq:quasi-dynamics}, we first formulate the contact forces resulting from the virtual displacement $\nu$, then project these forces on the system Jacobian to filter out unfeasible motions that may lie in the null-space of the Jacobian pseudo-inverse $(J {J}^\top)^{-1} J$. 
Following this yields the actual new state of the system $q^+$.
\begin{subequations}
\begin{align}
    f &= (J {J}^\top)^{-1} J (\mathbf{M} v - \mathbf{K} u) \\
    q^+ &= q + \mathbf{M}^{-1} (\mathbf{K} u + J^\top f).
\end{align}
\label{eq:dynamics_optimization}%
\end{subequations}

The objective function of the manipulation optimization is
\begin{subequations}
\begin{align}
    \min_{u, \nu} \quad & \alpha \mathcal{G}_p(q^{N_{cr}}) + \alpha \mathcal{G}_r(q^{N_{cr}}) + \mathcal{G}_d(q^{N_{cr}}) \\
    \text{with} \quad 
        \mathcal{G}_d(q) &= (q - q^{goal})^\top W_d (q - q^{goal}),
\end{align}
\end{subequations}
where $\alpha$ is a diagonal matrix of binary values encoding which contacts all the possible contact pairs between the robot and the object surface patches are currently active (1 for active contact, 0 for inactive).

At each step, the manipulation needs to satisfy the constraints described in \cref{eq:force-coulomb}, \cref{eq:constraint_place} and the constraints
\begin{subequations}
\begin{align}
    d_k^r f_k^n &= 0 \\
    -u_{max} &\leq q_a^+ - q_a \leq u_{max} \\
    -\nu_{max} &\leq \nu \leq \nu_{max},
\end{align}
\label{eq:constraint_manip}
\end{subequations}
where $\nu_{max}$ is an arbitrary boundary for object and robot virtual displacement during one state transition step.

\section{EXPERIMENTS}
\label{section:experiment-results}

To highlight the interests of our contribution, we conduct three experiments:
(1) a comparison of planning performance between our planner and the state of the art \gls{wmm} sampling planner to demonstrate the advantage of using continuous optimization in \gls{wmm}, enabled by our proposed contact surface representation; 
(2) a comparison of the performance between our planner with our proposed cost design and with another baseline cost design, to confirm the benefit of our cost design for avoiding local minimum and improving time performance; 
(3) a transfer to humanoid robot hardware with complex shape to verify the accuracy of our representation. 

All experiments are conducted on a computer with 16 $\times$ Intel(R) Core i9-9900K CPU $@$ 3.60GHz and Nvidia GeForce RTX2080 GPU.
We use the constant settings $N_{cf}=5$, $N_{cr}=1$, $\beta_d=10$, $\beta_p=5$, $\beta_r=10$ and $\beta_j=0.1$, tuned empirically to give satisfying results across all considered tasks and robots in the experiments.

\subsection{Continuous Contact Optimization VS Sampling}
\label{subsection:comparative-study}

\subsubsection{Protocol} 
We compare the performance of our planner with the sampling planner described in \cite{pang2023globalplanningcontactrichmanipulation} as a baseline. 
The system considered in this comparison corresponds to the iiwa\_bimanual example provided in \cite{pang2023globalplanningcontactrichmanipulation}.
It features two KUKA iiwa 7 robots and a 0.14m radius cylinder object constrained to move on a plane.
We consider two scenarios consisting in a translation with re-orientation to an increasing angle.
The scenario 1 is a lateral motion and the scenario 2 is a backward motion, and the re-orientation angle varies from $0^\circ$ to $180^\circ$.
Exact initial and goal coordinates can be found in \cref{fig:experiment_comparison}.
For each planning, we measured the planning time average, minimum and maximum values and the success rate over 15 attempts. 
The Coulomb friction is set to 0.5 on all surfaces. 
The greed parameter of our global planner is set to 1 for our method and the greed of the \gls{rrt} used in the baseline planner is set to 0.5 (which gave the best results for the baseline).
The Baseline is configured to use the free solvers OSQP and NLOPT \cite{NLopt} on the Drake simulator \cite{drake} and our method uses CasADI \cite{Andersson2018CasADiAS} with the free solver IPOPT \cite{Wchter2006OnTI} wrapped in the package OpTaS \cite{Mower_2023}.

\subsubsection{Results} 
The results shown in \cref{fig:experiment_comparison} empirically demonstrate that our method has several advantages compared to the baseline. 
The overall planning time is 77\% faster in average across both scenarios, which confirms our assumption that guiding the planning search with continuous optimization is beneficial. 
The planning time of our planner is more consistent, with less than 10mn difference between the minimum and maximum values, except for the scenario 1 with $180^\circ$ re-orientation. 
The success rate of our planner is also higher, with 100\% success rate for most of the scenarios. 
Finally, our method implements all kino-dynamic constraints at each step of the planning, so no refinement of the final trajectory is required, unlike the baseline planner which requires a refinement that sometimes fails. 

\subsection{Contact Optimization evaluation}
\label{subsection:cost-study}

\begin{figure}[t]
    \centering
    \resizebox{\columnwidth}{!}{%
    \begin{tabular}{|l r||r|r| }
        \hline
            \multicolumn{2}{|c|}{\textbf{Planning time [mn]}}
            & \multicolumn{1}{|c|}{Scenario 3} 
            & \multicolumn{1}{|c|}{Scenario 4} \\ 
        \hline
        \multicolumn{2}{|l|}{$q_u^{init}$ [m, m, deg]}
            & \multicolumn{1}{|c|}{(0.50, -0.55, 0.00)} 
            & \multicolumn{1}{|c|}{(0.40, 0.00, 0.00)} \\ 
        \hline
        \multicolumn{2}{|l|}{$q_a^{init}$ [deg]}
            & \multicolumn{1}{|c|}{(-45, 0, 0, -45, 0, 0)} 
            & \multicolumn{1}{|c|}{(0, 0, 0, 0, 0, 0)} \\ 
        \hline
        \multicolumn{2}{|l|}{$q_u^{goal}$ [m, m, deg]}
            & \multicolumn{1}{|c|}{(0.50, 0.55, -90.0)} 
            & \multicolumn{1}{|c|}{(0.80, 0.00, 0.00)} \\ 
        \hline
        \hline
        OURS
            & $\text{success}^1$ & 100\% & 100\%\\
            & mean [mn]& \textbf{5.6} & \textbf{2.6} \\
            & min [mn]& 5.2 & 2.3 \\
            & max [mn]& 6.1 & 2.7 \\
        \hline
        Baseline
            & $\text{success}^1$ & 100\% & 20\% \\
            & mean [mn]& 15.2 & 27.9 \\
            & min [mn]& 8.2 & 12.1 \\
            & max [mn]& 21.4 & - \\
        \hline
    \end{tabular}
    }
    
    \centering
    \captionsetup{justification=justified,margin=0.0cm}
    \caption{Comparison of planning performance between our planner with our proposed cost (OURS) and a baseline cost.
    $^{(1)}$ Planning time exceeding 30mn is considered as failure.}
    \label{fig:experiment_costdesign}
\end{figure}

\subsubsection{Protocol}
In this experiment we assess the improvement of the contact optimization induced by our cost design. 
To evaluate this, we compare the planning performance of our planning pipeline with our cost design and with the baseline cost from \cite{jin2024complementarity}.
The baseline cost is defined as the sum of squares of distances from the object center to all potential contacts on the robot surface.
We evaluate our planning with both cost designs on two complex scenarios 3 and 4. 
The scenario 3 consists in moving a 0.14m cylinder from the left edge of the table to the right edge, the robot arms starting on the right side of the table. 
This requires the robot to manipulate the object with only one arm first and to pass it to the other arm. 
The scenario 4 is a forward translation of the object from a pushing position difficult to reach with the robot end-effectors, encouraging the use of body parts with lower \gls{dof}.
The greed parameter of our planner is set to 1 for our cost design and 0.5 for the baseline cost.
Each scenario is planned 10 times for statistical results.

\subsubsection{Results}
The results in \cref{fig:experiment_costdesign} show that our cost design helps keeping reasonable planning time ($<10$mn) even on the more complex scenarios. 
On the contrary, our planner with baseline costs requires more time to find a solution, and even fails at finding one within 30mn for the scenario 4, 80\% of the time. 
In contrast the planner with our cost design is able to create better contacts that allow the robot to move the object right away, without relying on randomness, improving the repeatability of the planning at no cost.

\subsection{Hardware transfer}
\label{subsection:exp-hardware-transfer}

\begin{figure}[t!]
    \centering
    \includegraphics[width=1.0\linewidth,trim={0 7cm 0 0},clip]{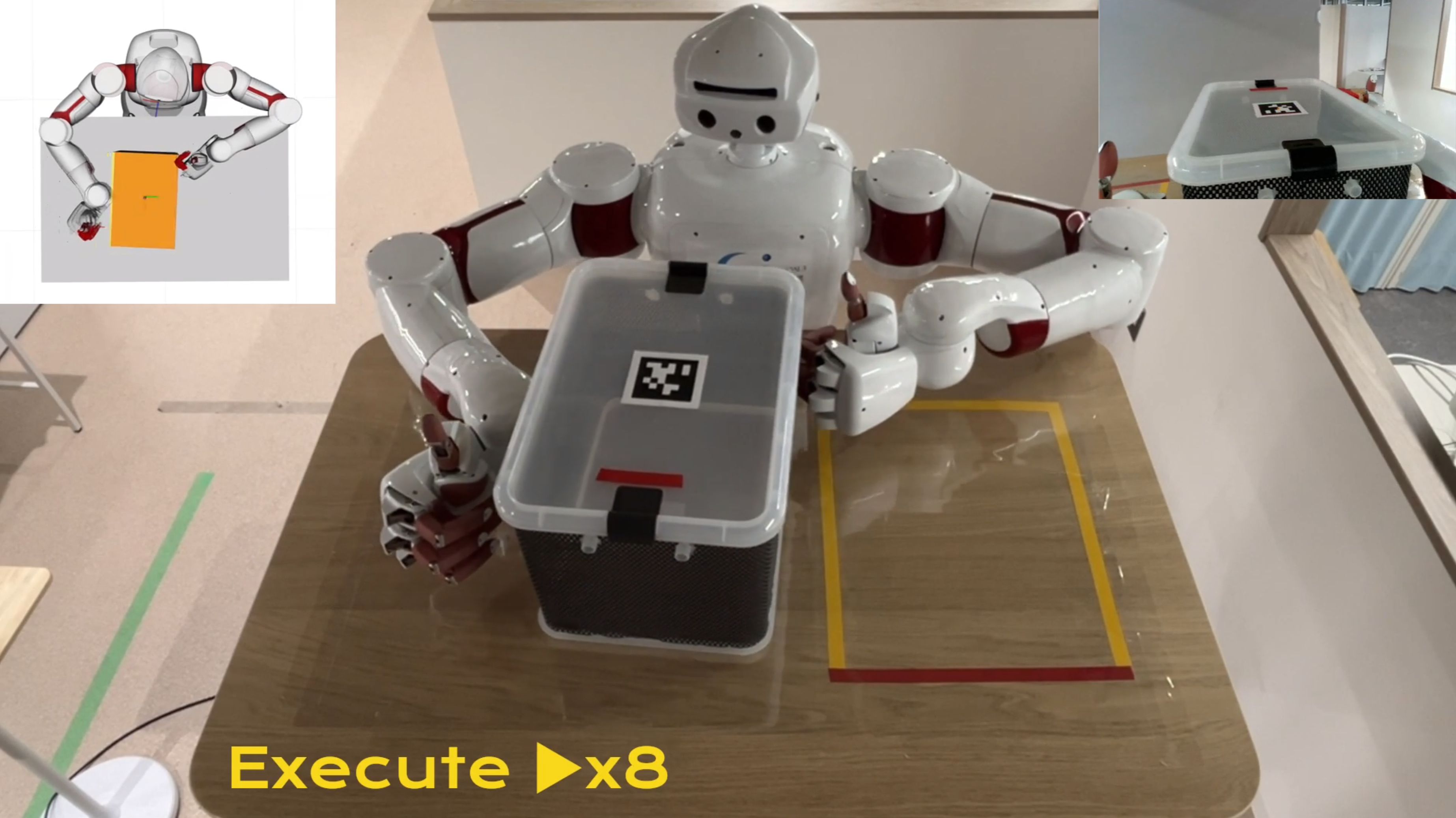}
    \caption{Experimental setup for hardware transfer on a humanoid robot platform with dual-arm and torso.}
    \label{fig:hardware_transfer}
\end{figure}

\subsubsection{Protocol}
In order to confirm the suitable accuracy of our proposed representation of the robot surface and the transferability of our planning to hardware, we conducted an experiment on a real robot with two arms and torso.
The robot has to manipulate a rounded rectangle box of two different sizes to a goal pose on the table using through contact with its two arms and torso.
We evaluate the three following specific scenarios: i) Move a small box (A4 size) from an arbitrary pose around $q_u^{init}=(0.54; -0.35; 90)$ to the goal pose $q_u^{goal}=(0.54; 0.29; 0)$ with a rotation of about $90^\circ$; ii) Move a large box (A3 size) from an arbitrary pose around $q_u^{init}=(0.55; -0.25; 0)$ to the same goal pose $q_u^{goal}=(0.54; 0.29; 0)$ without rotation; iii) Move the same large box from the pose $q_u^{init}=(0.54; 0.29; 0)$ to a goal pose near the torso $q_u^{goal}=(0.29; -0.13; 51.6)$, to encourage contact with the torso.
As we expect the friction of the actual system to differ from the model, and possibly be non-uniform, we allow re-planning manipulation when the box pose (position and orientation) deviates too much from the planned trajectory.
The box pose is tracked using an AprilTag \cite{5979561} placed on its top cover (see \cref{fig:hardware_transfer}).
We use an impedance control to control the robot.

\subsubsection{Results}
The result of this experiment is shown in the \href{https://levevictor.github.io/thelazyrobot/#humanoids2025}{video}. The experiment validates the accuracy of our representation by realizing planned contacts with many parts (fingers, wrist, elbow, torso) of a humanoid robot with complex non-convex geometry. It also demonstrate the convergence of our re-planning loop to the final solution, although the computation time in the order of 3 minutes for each re-planning is far from allowing us to integrate it in an \gls{mpc} as it is.

\section{DISCUSSION}
\label{section:discussion}

In this work we presented a new model to represent the robot and object surface with a closed-form solution for computing the two surfaces proximity points, enabling planning of \gls{wmm} using continuous \gls{nlp}. 
We also proposed a new cost function design to improve the contact placement on both the robot and object contact manifolds. These contributions together enabled the implementation of a framework for planning efficient planar \gls{wmm}: 
our method exhibits a planning time 77$\%$ faster than the state-of-the-art in average, which confirms the benefit of using continuous optimization over sampling approach for \gls{wmm}. 

Nonetheless, this approach is still facing numerous limitations to discuss. 
First, although we only validated our method on planar \gls{wmm} at the moment, our representation of segment to segment proximity can be easily adapted to 3D. 
Adapting the polygon proximity to suitable representation for 3D whole-body manipulation is in our future plans. 
Our cost definition in \cref{eq:contact_transition} extends to 3D, albeit we are yet to study its computational scalability to additional contact pairs.
Our method can theoretically adapt to non-convex shapes and articulated objects by using multiple patches, but such application is still unverified.
Second, our planning remains sensitive to local minima because of the non-convex nature of the problem. 
Better metrics to anticipate the horizon outcome of the manipulation could improve the performance in most complex scenarios. 
FInally, our optimization is too slow to be implemented in a \gls{mpc} for real-time \gls{wmm} tasks. Studying means for reaching real-time performance while preserving the accuracy of our method is within our next priorities.

\addtolength{\textheight}{-0cm}   






\bibliographystyle{IEEEtran} 
\bibliography{references} 

\end{document}